\documentclass[cameraready]{Interspeech}

\usepackage{tikz}
\usetikzlibrary{arrows.meta, positioning, fit, backgrounds}
\usepackage{pifont}    % \ding{51} checkmark, \ding{55} cross

\title{The Eloquence submission for Task 2 of the Interspeech 2026 \\MLC-SLM challenge}

\author[affiliation={1}, equalcontribution]{Jordi}{Luque}
\author[affiliation={2}, equalcontribution]{Lorenzo}{Concina}
\author[affiliation={2}]{Marco}{Matassoni}
\author[affiliation={2}]{Alessio}{Brutti}
\author[affiliation={3}]{Filippo}{Vella}

\address{
    $^1$ Telefónica Innovación Digital, Scientific Group \\
    $^2$ Fondazione Bruno Kessler \\
    $^3$ Consiglio Nazionale delle Ricerche
}

\email{ jordi.luque@telefonica.com, \{lconcina, brutti, matasso\}@fbk.eu,
 filippo.vella@icar.cnr.it}


\newcommand{\pos}[1]{\textcolor{red}{+#1}}   % WER increase = worse
\newcommand{\negd}[1]{\textcolor{teal}{$-$#1}} % WER decrease = better
\usepackage{comment}

\begin{document}

\maketitle

\begin{abstract}
This paper details the Eloquence team's approach to Task 2 of the 2nd MLC-SLM challenge at Interspeech 2026, which involves multilingual Multiple-Choice Question Answering (MCQA) across 21 languages. Three approaches are explored. First, we fine-tune Voxtral-Mini-3B via LoRA with cross-lingual data augmentation, ASR transcript augmentation and timestamp-aware audio cropping, achieving 0.72 macro-accuracy on evaluation Phase 2. Second, we apply multimodal in-context learning (ICL) to the frozen Voxtral-24B model to correct a strong label bias, reaching 0.81, our best result. Third, a training-free retrieval system based on a three-layer voice-anchored memory combining acoustic identity, semantic content, and a knowledge graph achieves 0.68. All three systems substantially outperform the official baseline.

\end{abstract}

\section{Introduction}

Recent progress in Large Language Models (LLMs) has driven a broad shift toward Speech LLMs that unify speech perception and language understanding within a single model, moving beyond transcription-centric pipelines ~\cite{Zheng2026BalancingAA,Shi2025TrainSI} toward systems that can reason over both the acoustic and semantic content of spoken interaction~\cite{surveySLM, SLM, mlcslm1Eloquence, wang-etal-2026-closing, Wang2025MMSUAM, Peng2026AUA}. Real-world conversational speech~\cite{Kankanala2025BenchmarkingHA,Wang2025MSUBenchTU}, however, differs substantially from read or single-speaker recordings: it contains overlapping turns, disfluencies, and multiple speakers, and this complexity is compounded in multilingual settings where training resources are unevenly distributed across languages~\cite{SpeechLLMsLowResourceScenarios}. The Multilingual Conversational Speech Language Model (MLC-SLM) Challenge~\cite{MLC2025} was introduced to address this gap by releasing large-scale, real-world multilingual conversational speech data together with a shared evaluation protocol. The first edition showed that current Speech LLMs largely solve transcription accuracy, while speaker diarization and deeper conversational understanding remained open; building on these findings, the second edition broadens language coverage and shifts emphasis toward diarization, acoustic understanding, and semantic understanding of full conversations\footnote{https://www.nexdata.ai/competition/mlc-slm}.

This paper reports on the participation of the Eloquence team in Task~2 of the second MLC-SLM challenge, \emph{Multilingual Conversational Speech Understanding}. Task~2 requires systems to answer multiple-choice questions probing both the acoustic properties (e.g.\ speaker traits, emotion) and the semantic content of an entire multi-speaker conversation~\cite{Wang2025IncorporatingCP,Maben2025AURAAF}. Crucially, no oracle information is available at evaluation time: no pre-segmented utterances, no speaker labels, and no ground-truth diarization are provided, so a complete system must resolve who is speaking, what is said, and how it is said, directly from the raw recording. The task leaves the choice of architecture open, encouraging both pipeline-based and end-to-end solutions, and evaluates systems on their ability to answer questions about the conversation as a whole.
We submitted three systems for Task~2, addressing the problem from different angles. The first, described in Section~\ref{sec:jordi}, fine-tunes Voxtral-Mini-3B via LoRA with cross-lingual data augmentation, ASR transcript augmentation, and timestamp-aware audio cropping. The second, described in Section~\ref{sec:ic24b}, applies in-context learning to the frozen Voxtral-24B speech-LLM to correct a strong label bias without any parameter update. The third, described in Section~\ref{sec:lorenzo}, reframes the task as a retrieval problem: rather than adapting a Speech-LLM through training, it equips a frozen LLM with a persistent, voice-anchored memory layer queried at inference time. 

%The remainder of the paper is organized as follows: Section~2 summarizes the challenge and the official baseline; Sections~3--4 describe our three submissions; Section~5 reports results; Section~6 concludes.
\begin{figure}
    \centering
    \includegraphics[width=1.0\linewidth]
    {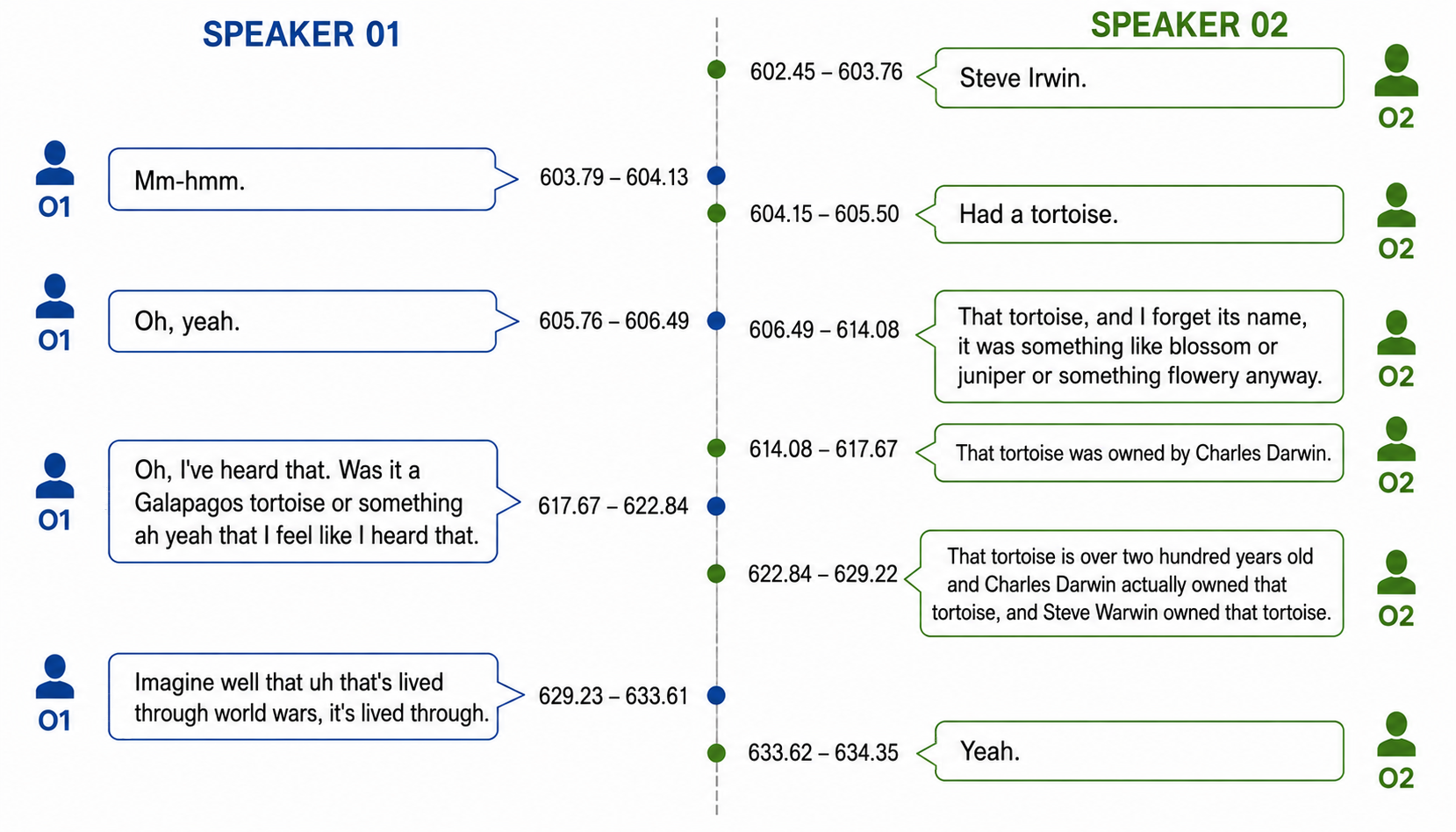}
    %{imgs/Chat_Australian_P05.png}
    \caption{Excerpt of 1141\_006 Australian English dialogue. }
    \label{fig:Australian_Dialogue}
\end{figure}
\section{Challenge and baseline description}
The MLC-SLM challenge is focused on LLMs and their adaptation capability to different languages and contexts. The remarkable results shown by LLMs in many tasks can be replicated with difficulty in a specific context or in multilingual settings, where some resources can be scarce and difficult to find. The challenge provides a real-world multilingual conversational speech dataset, see Fig.~\ref{fig:Australian_Dialogue}, with the aim of paving the ground for speech models that can respond to human counterpart naturally in multilingual, dynamic and context-rich environments. The dataset comprises diverse conversational styles and captures the complexities of human dialog, including pauses, interruptions, and speaker overlaps. The first challenge comprised 11 languages, including five regional varieties of English~\cite{MLC2025}. The second MLC-SLM challenge, which organizers describe as more challenging than the first, expanded the language coverage by adding Tagalog, Urdu, and Turkish, as well as a regional dialects of Canadian French, Mexican Spanish, and Brasilian Portoguese.
Task 1 is focused on multilingual conversational speech diarization and recognition, whereas Task 2, the task addressed by our system, is focused on multilingual conversational speech understanding. 
The training and development datasets provided are constructed using Gemini2.5-Pro, a multilingual multiple-choice question task that involves acoustic and semantic comprehension for training and development sets. The organizers provided a baseline system through Github\footnote{\url{https://github.com/DontPushMeee/MLC-SLM-2nd-Task2-Baseline}}, where Qwen2.5-Omni-7B is fine tuned using the ms-swift toolkit~\cite{zhao2024swiftascalablelightweightinfrastructure}.
The evaluation set of the challenge has been built as multiple-choice questions of speech comprehension with a similar procedure, and an additional manual review is used to rank the tasks.

\section{Fine-Tuning Speech-LLM}
\label{sec:jordi}

\subsection{Model and Training Setup}

We fine-tune Voxtral-Mini-3B-2507~\cite{voxtral_long} using LoRA~\cite{lora} ($r{=}16$, $\alpha{=}32$, all linear modules) on 4$\times$H100-64\,GB GPUs (effective batch 32, learning rate $2{\times}10^{-5}$ with linear decay). Each instance is a single-turn conversation placing audio and MCQ text in the user turn; the model emits a single letter (A--D) corresponding to the answer.
Since all provided audio files appear in the challenge evaluation, extended fine-tuning leads to session memorisation. We restricted training to $0.25\times$ an epoch, selecting the checkpoint minimizing the generalisation gap before memorisation sets in.

\subsection{Cross-Lingual Data Augmentation}
\label{ssec:augment}

Using \texttt{langdetect}~\cite{langdetect_python} we estimate that the evaluation set contains ${\approx}$52\% cross-lingual question pairs (non-English audio, English question), while training data have 0\% coverage. To account for this mismatch we translated non-English questions and options to English via NLLB-200~\cite{nllb200} (masking audio quotations, keeping native language terms for multilingual semantic questions) and train on a \emph{mixed} dataset of original + translated copies (6,630 questions for 124 files, keeping 25 files for validation).

\subsection{ASR Transcript Augmentation and Audio Cropping}
\label{ssec:crop}

We generate transcripts by fine-tuning Voxtral-Mini-3B as a LoRA ASR model: a Seed on MLC25 data, then an Adapted continuation with oversampled pre-training for the six new MLC26 languages. The Adapted model drops MLC26-new WER from 26.2\% to 15.2\%, with dramatic improvements on new languages (Table~\ref{tab:wer}). Audio sessions are chunked into 1-minute segments with loop-detection to mitigate hallucinations. Transcripts are prepended as \texttt{"Transcript: \{text\}"} to augment the MCQ prompt.
\begin{table}[t]
  \caption{WER (\%) Seed vs.\ Adapted Voxtral ASR. $\Delta$ = Adapted$-$Seed. MLC25: official test set; MLC26 ($^\dagger$): development set.}
  \label{tab:wer}
  \centering
%  \ttsize
  \setlength{\tabcolsep}{3pt}
  \begin{tabular}{lrrr}
    \toprule
    Language & Seed & Adapted & $\Delta$ \\
    \midrule
    English            &  7.0 &  6.9 & \negd{0.1} \\
    Spanish            &  8.2 &  7.9 & \negd{0.3} \\
    Vietnamese         &  9.1 &  9.0 & \negd{0.1} \\
    Thai               &  9.9 &  9.5 & \negd{0.4} \\
    Korean             &  8.8 & 10.1 & \pos{1.3} \\
    Portuguese         & 20.2 & 18.6 & \negd{1.6} \\
    Portuguese(Brazilian)$^\dagger$ & 18.4 & 10.9 & \negd{7.5} \\
    Russian            & 12.8 & 13.1 & \pos{0.3} \\
    Urdu$^\dagger$     &140.1 & 13.0 & \negd{127.1} \\
    French             & 15.2 & 15.3 & \pos{0.1} \\
    French(Canadian)$^\dagger$       & 34.7 & 18.3 & \negd{16.4} \\
    German             & 17.5 & 17.3 & \negd{0.2} \\
    Tagalog$^\dagger$  & 49.7 & 20.6 & \negd{29.1} \\
    Italian            & 18.6 & 20.1 & \pos{1.5} \\
    Turkish$^\dagger$  & 88.8 & 21.2 & \negd{67.6} \\
    Japanese           & 19.3 & 21.5 & \pos{2.2} \\
    \midrule
    \textbf{MLC25 Overall} & \textbf{11.8} & \textbf{12.0} & \pos{0.2} \\
    \textbf{MLC26 Overall}$^\dagger$ & \textbf{26.2} & \textbf{15.2} & \negd{11.0} \\
    \bottomrule
  \end{tabular}
\end{table}
Approximately 51\% of the questions in the development set include timestamp references pointing to the audio excerpts containing the answers. To improve training and inference efficiency and to encourage the Speech-LLM to focus on specific audio content, we cropped the audio to a $\pm 30$s window around these automatically detected timestamps. This approach yielded our best Phase 1 score of 0.85.
An ablation of the fine-tuning components is shown in Table~\ref{tab:ablation} reporting results submitted in Phase 1. 
 
\begin{table}[t]
  \caption{Ablation of fine-tuning components on Phase~1.}
  \label{tab:ablation}
  \centering
  \small
  \begin{tabular}{lcccc}
    \toprule
    Training Data &  Transc. & Crop & Acc.\\
    \midrule
    3B Baseline & No & No & 0.7162 \\
    + NLLB-only & No & No & 0.7382 \\
    + Mixed & Yes & No & 0.7283 \\
    + Mixed &  Yes & 30s & \textbf{0.7446} \\
    \bottomrule
  \end{tabular}
\end{table}

\section{In-Context Learning with Speech-LLM}
\label{sec:ic24b}

We detected a systematic class-A prediction bias that affects both Voxtral model sizes. The 24B baseline model predicts class~A for 63\% of questions and class~D for only 2\%.

\subsection{In-Context Learning}
\label{ssec:icl}

Prepending solved MCQ examples before the target question, i.e. few-shot In-Context Learning (ICL) corrects the bias in the 24B model for the English-audio/English-questions. Two text-only shots already shift class-A predictions from 63\% to 44\% and improve accuracy by $+$5.1\,pp; scaling to six multimodal shots adds a further $+$1.5\,pp, yielding our best score of \textbf{0.8095} (Table~\ref{tab:class_dist}). Note that the same set of ICL examples is used for all evaluation instances, consisting exclusively of English examples extracted from the training set. We hypothesize that employing multilingual language-specific ICL strategies could further improve performance. Each multimodal shot places a 30\,s audio clip in the user turn with an audio-focus prefix and MCQ text; the assistant turn is the bare answer letter. For 4-option questions, four shots cover one instance of each answer label (A, B, C, D) drawn from held-out audio, ensuring the model observes all four labels as valid outputs. For 2-option (A/B) questions, two shots are used.

\begin{table}[t]
  \caption{Predicted label distributions and Phase~2 accuracy (9,470 questions).
           All system use 30\,s crop in inference and no ASR transcripts as augmented context. mm stands for multimodal shots.}
  \label{tab:class_dist}
  \centering
  \small
  \begin{tabular}{lrrrrr}
    \toprule
    System & A & B & C & D & Acc.\\
    \midrule
    24B Baseline              & 63\% & 26\% &  9\% & 2\% & 0.743 \\
    24B + ICL-2 text      & 44\% & 37\% & 14\% & 6\% & 0.795 \\
    24B + ICL-6 mm        & 43\% & 37\% & 14\% & 6\% & \textbf{0.810} \\
    24B + ICL-2 + calib.  & 43\% & 37\% & 14\% & 6\% & 0.795 \\
    3B Baseline & 65\% & 24\% & 8\% & 2\% & 0.701  \\
    3B fine-tuned & 57\% & 32\% & 10\% & 1\% & 0.724  \\
    \bottomrule
  \end{tabular}
\end{table}

\subsection{Inference-Time Calibration}
We also evaluated an inference-time calibration strategy known as prior calibration (CBU, \cite{zhao2021calibrate}), which subtracts null-prompt log-probabilities to eliminate positional priors. When applied to ICL-2, this method reached the same accuracy as the baseline ICL-2 ($\approx$0.795), suggesting that in-context learning alone is sufficient to correct positional bias for the 24B model

%==================================================================================
\section{Training-Free Retrieval Approach}
\label{sec:lorenzo}

\subsection{A voice-anchored memory layer}
\label{ssec:system}

Our third approach treats Task~2 not as a model-training problem but as a \emph{retrieval} problem. We use a persistent voice-anchored memory layer that gives a standard LLM cross-session memory, semantic content memory, and a knowledge graph of facts and relationships~\cite{Wils2026SmalltalkKGAK}, without training any component. We enroll the speech session in our system and then at query time we retrieve from it the relevant context needed to answer the question. The enrollment of audio files in this memory layer is accomplished by an enrollment pipeline which segments a multispeaker conversation by utterance through diarization, then a single Speech-LLM Voxtral~Mini~3B~\cite{voxtral_long} served via vLLM\cite{vLLM_long}, performs the entire audio frontend: one call per utterance returns the transcription, language, gender, age, and both the acoustic and textual emotion labels. All of these extracted information are structured and stored in memory. The layer is then queried at runtime to assemble a structured \emph{fact sheet} that is handed to a frozen, swappable downstream LLM and helps it answering the question based on the retrived information; the entire system is composed of pre-trained models at inference time. This approach allows us to enroll the multi-speaker conversation once and then answer all the related questions efficiently without the need to process the audio file multiple times. For these experiments, we used LLM Qwen3-14B-AWQ~\footnote{\url{https://huggingface.co/Qwen/Qwen3-14B-AWQ}}.

\begin{figure*}[t]
\centering
\begin{tikzpicture}[
  scale=0.75,
  transform shape,
  font=\footnotesize,
  >={Stealth[length=2mm,width=2mm]},
  layer/.style={
    rectangle, rounded corners=2pt, draw=black!70, thick,
    minimum width=3.6cm, minimum height=1.5cm,
    align=center, fill=blue!5
  },
  ingest/.style={
    rectangle, rounded corners=2pt, draw=black!50,
    minimum width=2.2cm, minimum height=0.7cm,
    align=center, fill=gray!10, font=\scriptsize
  },
  output/.style={
    rectangle, rounded corners=2pt, draw=black!50,
    minimum width=2.2cm, minimum height=0.7cm,
    align=center, fill=orange!15, font=\scriptsize
  },
  query/.style={
    rectangle, rounded corners=2pt, draw=black!50,
    minimum width=2.2cm, minimum height=0.7cm,
    align=center, fill=yellow!15, font=\scriptsize
  },
  retrieval/.style={
    rectangle, rounded corners=2pt, draw=black!70, thick,
    minimum width=2.0cm, minimum height=1.0cm,
    align=center, font=\scriptsize, fill=cyan!8
  },
  uuid/.style={
    rectangle, draw=red!60, thick, fill=red!5,
    minimum width=0.9cm, minimum height=0.5cm,
    font=\scriptsize\ttfamily
  },
  emoTag/.style={
    rectangle, rounded corners=1pt, draw=purple!50, thick,
    fill=purple!8, font=\scriptsize, inner sep=2pt
  },
  arr/.style={->, thick, black!70},
  arrr/.style={->, thick, black!70, dashed}
]
\node[ingest] (audio) at (0, 1.5) {Audio recording\\\scriptsize(diarization $\to$ segments)};

\node[ingest] (chat)  at (0, -0.2)   {Chat session\\\scriptsize(user/assistant turns)};

\node[uuid] (uuid) at (3.2, 0.50) {UUID $u$};

\node[layer] (ac)  at (7.5, 2.4)  {\textbf{$L_{\mathrm{ac}}$: Acoustic Identity}\\TitaNet 192-d embedding\\ChromaDB };

\node[layer] (sem) at (7.5, 0.5)  {\textbf{$L_{\mathrm{sem}}$: Semantic Content}\\MiniLM 384-d embedding\\ChromaDB };
\node[layer] (kg)  at (7.5, -1.4) {\textbf{$L_{\mathrm{kg}}$: Knowledge Graph}\\NetworkX, typed FACTs\\};
\node[emoTag] (emoSem) at (9.4, 1.25) {emotion meta};
\node[emoTag] (emoKg)  at (9.4, -0.65) {emotion meta};

\node[query]      (question)  at (12.5, 2.0)  {\textbf{User question}\\\scriptsize (+ optional probe audio)};
\node[retrieval]  (retrieval) at (12.5, 0.5)  {\textbf{Retrieval}};
\node[output]     (factsheet) at (12.5, -0.8) {\textbf{Fact sheet}\\\scriptsize speaker + content\\\scriptsize + facts + emotion};
\node[output]     (answerllm) at (12.5, -2.0) {\textbf{LLM}};
\node[output]     (answer)    at (12.5, -3.1) { Answer};

\draw[arr] (audio.east) -- (uuid.west);
\draw[arr] (chat.east)  -- (uuid.west);
\draw[arr] (uuid.east) to[out=25,in=180]  (ac.west);
\draw[arr] (uuid.east) --                  (sem.west);
\draw[arr] (uuid.east) to[out=-25,in=180] (kg.west);
\draw[arr] (question.south) -- (retrieval.north);
\draw[arrr] (retrieval.west) to[out=180,in=0] (ac.east);
\draw[arrr] (retrieval.west) -- (sem.east);
\draw[arrr] (retrieval.west) to[out=180,in=0] (kg.east);
\draw[arr]  (retrieval.south) -- (factsheet.north);
\draw[arr]  (factsheet.south) -- (answerllm.north);
\draw[arr]  (answerllm.south) -- (answer.north);
\node[font=\scriptsize, anchor=west] at (0, -3.8) {
  \tikz \draw[arr] (0,0) -- (0.5,0);\ enrollment write / question flow
  \quad
  \tikz \draw[arrr] (0,0) -- (0.5,0);\ retrieval read
  \quad
  \tikz \node[emoTag, inner sep=1.5pt] {meta};\ cross-layer attribute
};
\end{tikzpicture}
\caption{System architecture. Every utterance receives a UUID $u$ at ingestion. Audio inputs are diarized into segments and enrolled into the three layers, while chat sessions skip the acoustic layer. At query time, the Retrieval component receives a user question (optionally with a probe audio clip), reads from each layer (dashed lines), and assembles a fact sheet that is handed to a swappable answer LLM.}
\label{fig:architecture}
\end{figure*}
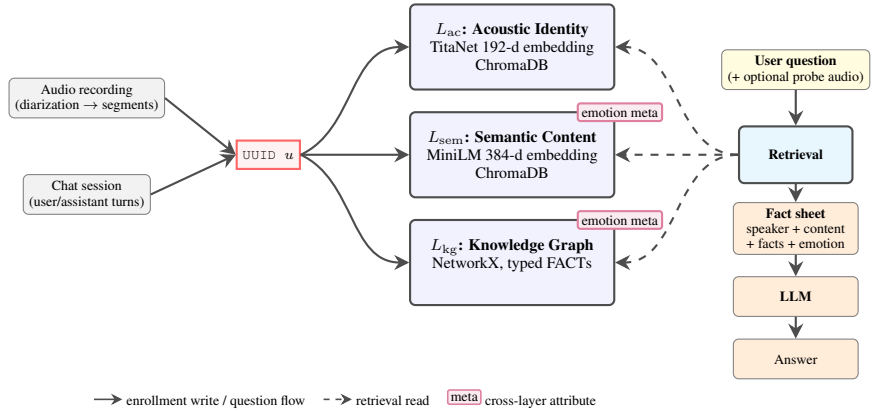

\textbf{Shared utterance identifiers.}
 Memory is organized as three persistent storage layers, illustrated in Figure~\ref{fig:architecture}: an acoustic-identity layer $L_{\mathrm{ac}}$, a semantic-content layer $L_{\mathrm{sem}}$, and a knowledge-graph layer $L_{\mathrm{kg}}$. Every turn ingested by the system --- whether produced by diarization of a multi-speaker recording or by a chat turn --- is assigned a single UUID $u$ at ingestion, which is propagated as the primary key across the layers that record it. For audio utterances $u$ is generated when the speaker embedding is written to $L_{\mathrm{ac}}$ and reused as the record ID in
$L_{\mathrm{sem}}$ and as the utterance-node identifier in $L_{\mathrm{kg}}$; chat turns skip $L_{\mathrm{ac}}$ and originate $u$ in $L_{\mathrm{sem}}$. The invariant is that whenever two layers store information about the same utterance, they store it under the same key, so cross-layer references resolve by direct lookup rather than approximate matching.

\textbf{Acoustic identity ($L_{\mathrm{ac}}$).}
For each utterance, a 192-dimensional speaker embedding is extracted with
Titanet-Large~\cite{titanet_long} and stored in a persistent ChromaDB collection.
Records are tagged as \emph{enroll} data (durable identity profiles for known
speakers) or \emph{probe} data (transient embeddings produced during a single
query session). At query time, the layer supports \emph{identification}
(comparing a probe to enrolled profiles by top-$k$ cosine retrieval and
returning the closest identity above a similarity threshold) and
\emph{resolution} (matching a probe against other probes from the same session
so that segments of the same unknown speaker are linked).

\textbf{Semantic content ($L_{\mathrm{sem}}$).}
Utterance text---transcribed by Voxtral for audio and taken verbatim for chat---is encoded with all-MiniLM-L6-v2~\cite{minilm_long} and stored in a parallel
ChromaDB collection under the same UUIDs as $L_{\mathrm{ac}}$. Per-record
metadata supports speaker-conditioned semantic search: a textual query can be restricted to utterances produced by one or more identified speakers, optionally
filtered further by emotion, time, or dialogue membership.

\textbf{Knowledge graph ($L_{\mathrm{kg}}$).}
The knowledge graph layer answers \emph{how facts relate and how they evolve}. It is a directed graph in NetworkX~\cite{networkx_long} with four types of nodes (speaker, utterance, entity, and dialogue) and two classes of edges. We use a local Qwen3-14B-AWQ served by vLLM as a local extractor to extract nodes and relationships from conversation transcription.

\emph{Structural edges} are added eagerly when an utterance is ingested and require no LLM call:
\begin{itemize}
    \item \texttt{SAID} (speaker $\rightarrow$ utterance)
    \item \texttt{CONTAINS} (dialogue $\rightarrow$ utterance)
    \item \texttt{HAS\_PARTICIPANT} (dialogue $\rightarrow$ speaker)
    \item \texttt{KNOWS} (speaker $\leftrightarrow$ speaker), with co-occurrence counts accumulated across dialogues
\end{itemize}

\emph{Semantic edges} computed by an LLM extractor:
\begin{itemize}
    \item \texttt{MENTIONS}: from utterance to entity
    \item \texttt{FACT}: typed edges between entities
\end{itemize}

 Each fact carries provenance and two timestamps, \emph{valid-from} and \emph{valid-to}: a new fact that contradicts an existing one does not delete it, but closes its validity interval, preserving history. Newly extracted entities are reconciled against the graph by an entity resolver that embeds each candidate with the same sentence-transformer used in $L_{\mathrm{sem}}$ and searches a dedicated entity collection, auto-merging above a high similarity
threshold, treating low-similarity candidates as new, and deferring ambiguous cases to an LLM judge.

\textbf{Emotion as cross-layer metadata.}
Each utterance is annotated with two emotion labels attached as metadata to $L_{\mathrm{sem}}$ and $L_{\mathrm{kg}}$: an acoustic label from prosody and a textual label from lexical content, both produced by Voxtral in the same call as the transcription. The two channels are kept independent rather than fused, since
tone and words can disagree and both carry signal; acoustic emotion is computed only for audio utterances.

\textbf{Fact sheet assembly.}
At query time, the retrieval pipeline reads the three layers and assembles a
fact sheet. Given an optional probe clip, $L_{\mathrm{ac}}$ identifies the
speaker against the enrolled profiles; $L_{\mathrm{sem}}$ is queried with the
question, filtered by the identified speaker(s); and $L_{\mathrm{kg}}$ is read in
one of three modes---profile, dialogue, or entity---selected by a lightweight
question classifier, returning emotion counts, participant lists, or
currently-valid facts respectively. The assembled fact sheet---per-utterance
entries from $L_{\mathrm{sem}}$, speaker identity from $L_{\mathrm{ac}}$, graph
context from $L_{\mathrm{kg}}$, and emotion labels---is the only context passed
to the downstream LLM, with a prompt that constrains the model to rely on
fact-sheet content alone.

\subsection{Application to the MLC-SLM Task~2}
\label{ssec:mlcslm}

%\textbf{Enrollment.} 
Sessions are processed one at a time. Each conversation is diarized with \texttt{pyannote.audio}~\cite{pyannote, Bredin23, Plaquet23}; every resulting segment is embedded with Titanet, transcribed and attribute-tagged by Voxtral, and written to all three layers ($L_{\mathrm{ac}}$, $L_{\mathrm{sem}}$, $L_{\mathrm{kg}}$) under a shared UUID. Diarized segments are clustered and matched against previously enrolled speakers, giving the system a cross-segment notion of \emph{who} is speaking that is anchored in voice rather than in an external speaker label.
%\textbf{Answering.} 
Task~2 is posed as multiple choice. For each question, the system performs speaker-conditioned semantic retrieval over the enrolled utterances of the current session and traverses the knowledge graph, then assembles the fact sheet described in Section~\ref{ssec:system}. The fact sheet, together with the question and its candidate options, is the only context given
to a frozen Qwen answer LLM, which is constrained to emit a single option letter. Because identity, content, and emotion are resolved by the retrieval layer before the LLM is invoked, the answer model performs neither speaker verification nor long-context recall itself.

\vspace{-0.25cm}
\section{Results}

Table~\ref{tab:results} summarizes the accuracy scores of all three developed systems.

\textbf{Fine-tuned Voxtral-3B.} The combination of translated data augmentation and 30s cropping yielded our highest development performance (0.85). While the mixed dataset (see Table~\ref{tab:ablation}) and timestamp-based cropping improved accuracy by directing the audio encoder toward relevant segments, the gains from transcription-based context augmentation and overall fine-tuning did not generalize to the evaluation sets. The resulting drop in score to 0.72 is likely attributable to overfitting or the LLM learning language-specific shortcuts during the fine-tuning process to a small dataset.

\textbf{Voxtral-24B with ICL.} The 24B model with six multimodal in-context shots reaches 0.81 on Phase~2, our best result overall. ICL corrects the strong label bias while requiring zero parameter updates or data for training. However, prior calibration provide no further gain over ICL, see Table~\ref{tab:class_dist}.

\textbf{Training-free retrieval.} We evaluated two variants differing only in the audio frontend: a multi-model pipeline (Whisper\cite{whisper} + Wav2Vec2 + RoBERTa) and a unified Voxtral~Mini~3B frontend. The unified frontend outperformed it on the dev set ($0.83$ vs.\ $0.78$) while simplifying the pipeline, and was submitted to the evaluation set where it reached $0.68$. This result is obtained without any training on the challenge corpus, indicating that a retrieval layer over frozen perception and language models is a competitive alternative to model adaptation, while remaining cheap to deploy and agnostic to the choice of downstream LLM.

\begin{table}[t]
  \caption{Accuracy scores on the MLC-SLM Task~2 development and Phase-2 test sets.}
  \label{tab:results}
  \centering
  \small
  \begin{tabular}{lcc}
    \toprule
    System & Dev & Test \\
    \midrule
    Official baseline & 0.35 & -- \\
    \midrule
    Voxtral-3B fine-tuned & 0.85 & 0.72 \\
    Voxtral-24B + ICL-6 & -- & 0.81 \\
    Train-free (multi-model) & 0.78 & -- \\
    Train-free (Voxtral frontend) & 0.83 & 0.68 \\
    \bottomrule
  \end{tabular}
\end{table}

\vspace{-0.2cm}
\section{Conclusions}

We presented three systems for Task 2 of the second MLC-SLM challenge under a fully blind evaluation setting, utilizing no oracle segmentation, speaker labels, or diarization during inference. Our fine-tuning experiments demonstrate that NLLB-200 translation effectively bridges the 52\% cross-lingual gap, and with timestamp-aware cropping achieved 0.85 on the development set; however it failed to generalize to the Phase~1 and Phase~2 evaluation sets due to overfitting. In contrast, applying in-context learning to the frozen Voxtral-24B model successfully mitigated strong pretrained label bias without parameter updates, achieving our best result of 0.81 on Phase 2. Finally, our training-free retrieval system (0.83 dev, 0.68 Phase~2) shows that a voice-anchored memory layer over frozen models is a competitive, low-cost alternative to task-specific adaptation. Together, these three approaches offer complementary trade-offs between accuracy, training cost, and deployment simplicity, placing the Eloquence team 5th in the final ranking.

\section{Acknowledgments}
This work has received funding from the European Union's Horizon Europe research and innovation programme under the project ELOQUENCE (Grant Agreement No. 101135916). This work was supported by computational resources from the EuroHPC Joint Undertaking under the EuroHPC AI Factory grant EHPC-AIF-2026LS01-004.

\bibliographystyle{IEEEtran}
\bibliography{mybib}

\end{document}